\documentclass[conference]{IEEEtran}

\IEEEoverridecommandlockouts

\usepackage{cite}
\usepackage[T1]{fontenc}
\usepackage{xcolor}
\usepackage{graphicx}
\usepackage{amsmath,amssymb,amsfonts}
\usepackage{pifont}
\usepackage[caption=false,font=footnotesize]{subfig}
\usepackage{multirow}
\usepackage{booktabs}
\usepackage{comicneue}
\usepackage{hyperref}

\hypersetup{
    colorlinks=true,
    citecolor=green,
    linkcolor=red,
    urlcolor=blue
}

\newcommand{\cmark}{\ding{51}}
\newcommand{\xmark}{\ding{55}}
\newcommand{\bluecheck}{{\color{blue}\cmark}}
\newcommand{\redcheck}{{\color{red}\xmark}}

\definecolor{c2fColor}{RGB}{157, 91, 82}
\definecolor{CoordAtt}{RGB}{118,144,60}
\definecolor{c3k2Color}{RGB}{56,122,153}
\definecolor{nms}{RGB}{153,153,153}

\begin{document}

\title{{\comicneue\textbf{MicroCharNet}}\\
Less is More for License Plate Character Detection}

\author{
\IEEEauthorblockN{
Huy Che\textsuperscript{1, 2}, 
Dinh-Duy Phan\textsuperscript{1, 2} and 
Duc-Lung Vu\textsuperscript{1, 2,*}
}
\IEEEauthorblockA{\textsuperscript{1}University of Information Technology, Ho Chi Minh City, Vietnam}
\IEEEauthorblockA{\textsuperscript{2}Vietnam National University, Ho Chi Minh City, Vietnam}
\IEEEauthorblockA{
\textsuperscript{*}Corresponding author: Duc-Lung Vu
}
Email: huycq@uit.edu.vn, duypd@uit.edu.vn, lungvd@uit.edu.vn
}

\maketitle              

\begin{abstract}
License plate character detection is a crucial component of intelligent transportation systems, where high accuracy and computational efficiency are required for real-time deployment. Although recent deep learning-based methods have substantially improved detection performance, many high-accuracy models rely on large-scale architectures that incur substantial computational overhead, limiting their applicability to resource-constrained devices. In this paper, we propose {\comicneue\textbf{MicroCharNet}}, an ultra-lightweight model specifically designed for license plate character detection. The proposed architecture employs a compact backbone composed of \textcolor{c2fColor}{\textbf{C2f}} blocks, integrated with \textcolor{CoordAtt}{\textbf{CoordAtt}} module to enhance feature extraction while preserving spatial information. A lightweight \textcolor{c3k2Color}{\textbf{C3k2}}-based neck fuses multi-level features, followed by a single-level anchor-free detection head that enables end-to-end prediction. Experiments conducted on the UFPR-ALPR dataset demonstrate that {\comicneue\textbf{MicroCharNet}} achieves competitive detection accuracy with only 0.08M parameters and 0.096 GFLOPs, while outperforming several recent YOLO-based baselines. Hardware-level evaluations further confirm its efficiency for real-time deployment on edge devices. These results indicate that carefully designed ultra-lightweight architectures can effectively balance accuracy and efficiency in license plate character detection. The source code is available at \hyperlink{https://github.com/chequanghuy/MicroCharNet}{https://github.com/chequanghuy/MicroCharNet}.
\end{abstract}


\section{Introduction} \label{intro}

License plate recognition is a critical component in many intelligent transportation systems, including vehicle surveillance \cite{surveillance}, smart parking \cite{parking}, and traffic violation management \cite{violation}. In practical deployment scenarios, particularly on edge cameras or resource-constrained computing platforms, recognition models are required not only to achieve high accuracy but also to ensure fast inference for stable real-time operation.

In recent years, deep learning models have significantly improved the performance of character and character-sequence recognition in natural images. However, most high-accuracy architectures are associated with a large number of parameters and high computational cost, making their direct deployment on resource-limited edge devices challenging. Among existing approaches, the YOLO family \cite{yolov8,yolov9,yolov10,yolov11,yolov12,yolo26} provides an effective solution for real-time license plate character detection, thanks to its fast, flexible object detection. Nevertheless, since these architectures were originally designed for general-purpose computer vision tasks such as object detection \cite{ctm,yolov9,yolov10}, image segmentation \cite{yolo_seg}, and multitask estimation \cite{yolo_multi1,yolo_multi2}, they are not truly optimized for the specific characteristics of license plate character detection. When applied directly, they often fail to fully exploit the fine-grained geometric features of characters, their sequential relationships, and the narrow regions of interest that are central to this task. As a result, lightweight networks trained on general-purpose vision tasks often do not achieve optimal performance when transferred to specialized application domains, such as license plate character detection \cite{ctm,parking,surveillance,violation}.

\begin{figure}
    \centering
    \includegraphics[width=\linewidth]{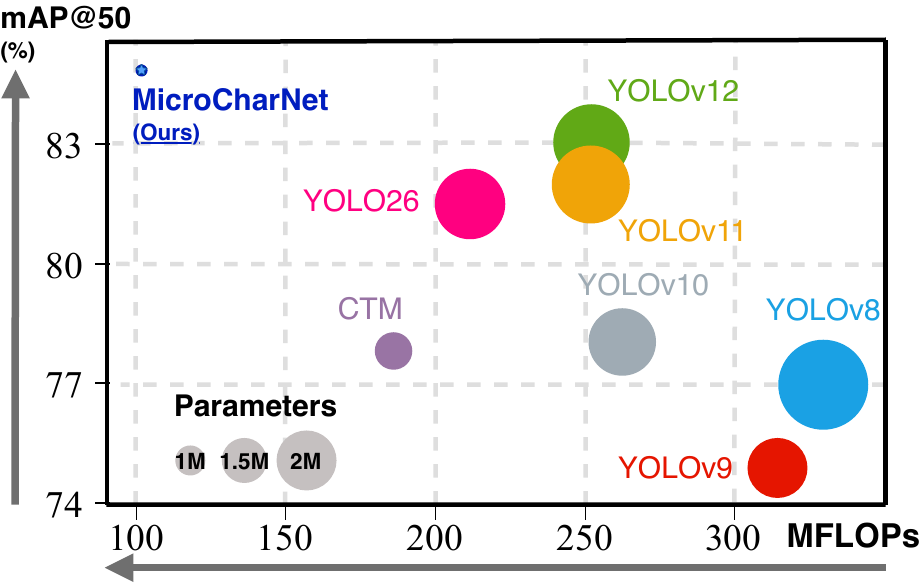}
    \caption{Accuracy–efficiency comparison between {\comicneue\textbf{MicroCharNet}} and YOLO-based baselines on the test set of UFPR-ALPR dataset.}
    \label{fig:highlight}
\end{figure}

Unlike approaches that prioritize large-scale models to maximize accuracy, this study focuses on a configuration suitable for practical deployment on resource-constrained devices. Our objective is not only to develop a compact model, but also to demonstrate that an architecture specifically tailored for license plate character recognition can maintain competitive performance while significantly reducing computational complexity and deployment cost. In particular, the proposed {\comicneue\textbf{MicroCharNet}} contains only 82.5K parameters and requires 0.096 GFLOPs, while still achieving competitive accuracy on the character detection task of the UFPR-ALPR dataset \cite{UFPR-ALPR}. These results indicate that ultra-lightweight models, when designed appropriately, have strong potential to simultaneously satisfy the requirements of accuracy, and deployability in real-world license plate recognition systems.

\begin{figure*}[!t]
    \centering
    \includegraphics[width=0.999\linewidth]{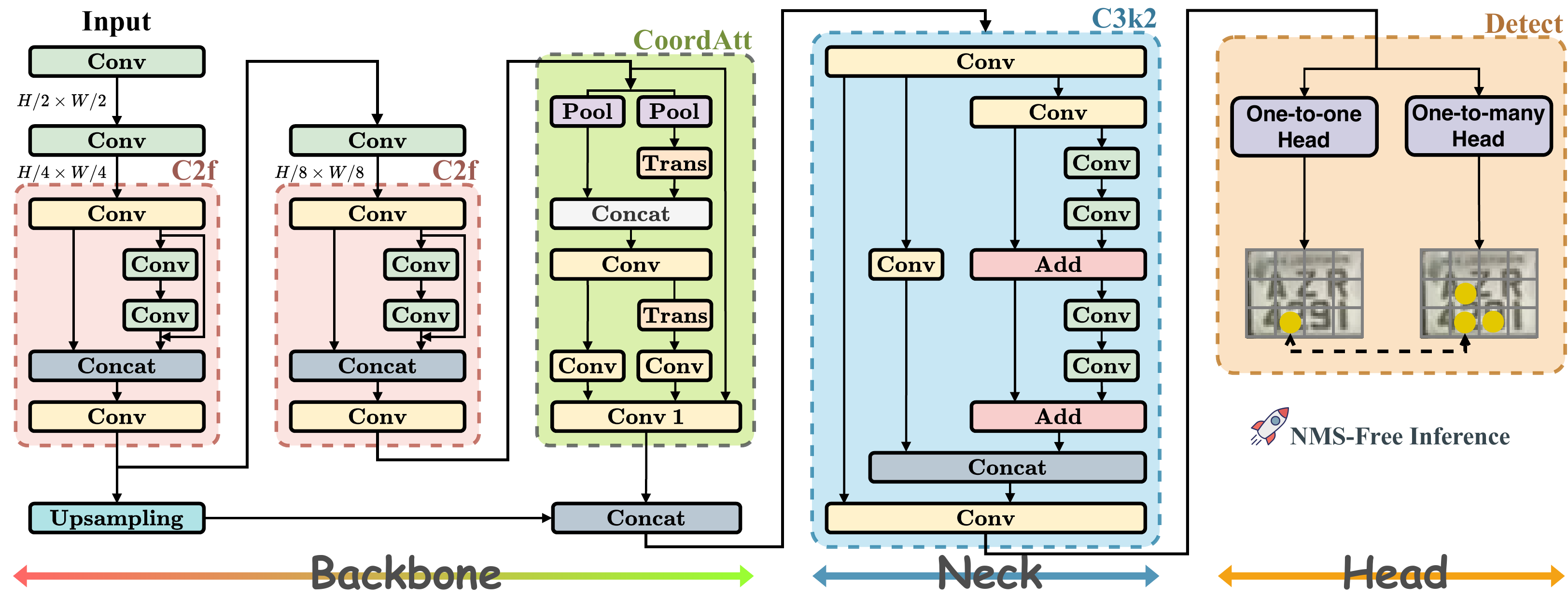}
    \caption{Overall architecture of {\comicneue\textbf{MicroCharNet}}, consisting of a lightweight \textcolor{c2fColor}{\textbf{C2f}}-based backbone, \textcolor{CoordAtt}{\textbf{CoordAtt}} module, \textcolor{c3k2Color}{\textbf{C3k2}}-based neck, and single-level anchor-free detection head.}
    \label{fig:model_}
\end{figure*}

The contributions of this paper are summarized as follows: (1) We propose {\comicneue\textbf{MicroCharNet}}, an ultra-lightweight architecture tailored to license plate character detection for real-time deployment on resource-constrained devices. (2) Extensive experiments on the UFPR-ALPR dataset \cite{UFPR-ALPR} demonstrate competitive performance with respect to both detection accuracy and computational efficiency. (3) Its practical deployment capability is further validated through inference-speed and runtime-efficiency evaluations on diverse hardware platforms.

\section{Related Work} \label{related}

Automatic license plate recognition is an important problem in computer vision and intelligent transportation systems. In addition to license plate localization, character prediction plays a particularly critical role, as it directly determines the final accuracy of the recognized license plate information. To address this task, a variety of approaches have been proposed, including object detection \cite{UFPR-ALPR,plate_detect,ctm}, Seq2Seq \cite{crnn}, and image segmentation \cite{plate_seg}. The rapid development of deep learning has promoted end-to-end and one-stage methods, enabling models to learn discriminative features directly from data and significantly improving performance on real-world license plate datasets. Among these approaches, object detection–based methods \cite{UFPR-ALPR,plate_detect,ctm} have attracted considerable attention due to their flexibility, predictive effectiveness, and practical ease of deployment. In recent years, the YOLO family \cite{yolov8,yolov9,yolov10,yolov11,yolov12,yolo26} has been widely adopted for real-time object detection due to its favorable balance between accuracy and inference speed. However, most YOLO architectures were originally designed for general-purpose vision tasks and are therefore not fully optimized for license plate character detection, where targets are typically small, narrow, and densely arranged. Moreover, highly accurate models are often associated with a large number of parameters and substantial computational cost, which limits their applicability on edge devices and resource-constrained embedded platforms.

A large of research has therefore focused on lightweight architectures \cite{twinplus,twinmixing,mobilenet,shufflenet} that reduce computational complexity while maintaining competitive performance. In parallel, attention modules \cite{cbam,coordatt,se,pcaa} have been introduced to enhance feature representation by enabling networks to emphasize informative regions and improve localization capability. Nevertheless, simply shrinking a general-purpose architecture or incorporating attention modules does not necessarily yield optimal performance for license plate character detection. This task requires not only the preservation of fine-grained local details but also the precise modeling of subtle spatial relationships among closely spaced characters. As a result, recent efforts have increasingly shifted toward lightweight models tailored to the target application domain.

Motivated by these observations, this study develops an ultra-lightweight architecture for license plate character detection that combines a compact design with enhanced spatial information modeling, aiming to achieve a favorable balance among detection accuracy, computational cost, and practical deployability across diverse hardware platforms.

\section{Method}

Our license plate character detection model is based on three components, as shown in detail in Figure \ref{fig:model_}.

\subsection{Backbone}

Given an input image $\mathbf{I} \in \mathbb{R}^{3 \times H \times W}$, the backbone first performs hierarchical feature extraction. In this stage, convolutional layers with a stride of 2 are employed to progressively reduce spatial resolution while increasing channel dimensionality, enabling the model to capture local patterns at different levels of abstraction efficiently. On the intermediate feature maps, \textcolor{c2fColor}{\textbf{C2f}} blocks are introduced after downsampling stages to enhance feature representation while maintaining a lightweight computational structure.

The \textcolor{c2fColor}{\textbf{C2f}} module follows a \textit{split--transform--concatenate} paradigm. Given an input feature map $X_{c2f} \in \mathbb{R}^{C_1 \times H \times W}$, it is first projected by a $1 \times 1$ convolution and then split along the channel dimension:
\begin{equation}
X_1, X_2 = \mathrm{split}(\mathrm{Conv}_{1 \times 1}(X_{c2f}))
\end{equation}

One branch is preserved to maintain the original information flow, while the other branch is refined through lightweight local transformations:
\begin{equation}
\hat{X}_2 = X_2+ \mathrm{Conv}_{3 \times 3}\big(\mathrm{Conv}_{3 \times 3}(X_2)\big)
\end{equation}
The features are then fused via concatenation followed by a $1 \times 1$ convolution:
\begin{equation}
Y_{c2f} = \mathrm{Conv}_{1 \times 1}\big(\mathrm{Concat}(X_1, \hat{X}_2)\big).
\end{equation}

This design improves feature reuse and information flow while reducing redundant transformations compared to heavier residual blocks. For license plate character detection, such a mechanism is particularly beneficial, as it preserves fine-grained edge details while learning stable geometric character patterns.

\begin{figure}
    \centering
    \includegraphics[width=\linewidth]{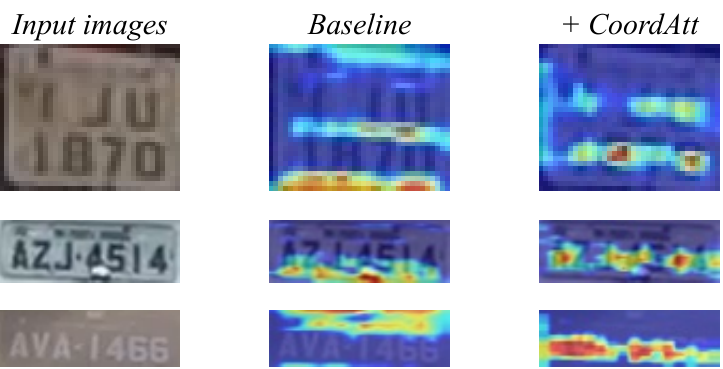}
    \caption{Visual representation of the class activation maps (CAM) before and after incorporating Coordinate Attention.}
    \label{fig:gradcam}
\end{figure}

\begin{table*}[!t]
\centering
\caption{Comparison of {\comicneue\textbf{MicroCharNet}} with state-of-the-art YOLO models.}
\label{tab:results}
\begin{tabular}{lccccccccc}
\toprule
\multirow{2}{*}{\textbf{Model}} & \multirow{2}{*}{\textbf{NMS Free}} & \multicolumn{2}{c}{\textbf{Validation set}} & \multicolumn{2}{c}{\textbf{Test set}} & \multirow{2}{*}{\textbf{\#Params}} & \multirow{2}{*}{\textbf{GFLOPs}} & \multirow{2}{*}{\textbf{\begin{tabular}[c]{@{}c@{}}Inference\\ (ms)\end{tabular}}} & \multirow{2}{*}{\textbf{\begin{tabular}[c]{@{}c@{}}Post-process\\ (ms)\end{tabular}}} \\ 
\cmidrule(lr){3-4} \cmidrule(lr){5-6}
& & \textbf{mAP@50} & \textbf{mAP@50-95} & \textbf{mAP@50} & \textbf{mAP@50-95} & & & & \\ 
\midrule
CTM \cite{ctm}   & \redcheck & 0.79 & 0.49 & 0.78 & 0.48 & \underline{1.26M} & 0.186 & 1.298 & 0.163 \\
YOLOv8 \cite{yolov8}       & \redcheck & 0.81 & \underline{0.51} & 0.77 & 0.48 & 3.01M & 0.325 & 1.465 & 0.171 \\
YOLOv9 \cite{yolov9}      & \redcheck & 0.77 & 0.48 & 0.75 & 0.45 & 1.98M & 0.306 & 2.443 & 0.163 \\
YOLOv10 \cite{yolov10}     & \bluecheck & 0.78 & 0.49 & 0.78 & 0.48 & 2.27M & 0.263 & 1.477 & \underline{0.023} \\
YOLOv11 \cite{yolov11}     & \redcheck & \underline{0.81} & \textbf{0.52} & 0.82 & \textbf{0.51} & 2.59M & 0.254 & 1.653 & 0.169 \\
YOLOv12 \cite{yolov12}     & \redcheck & 0.80 & 0.50 & \underline{0.83} & \textbf{0.51} & 2.56M & 0.254 & 1.990 & 0.176 \\
YOLO26 \cite{yolo26}      & \bluecheck & 0.80 & \underline{0.51} & 0.81 & \underline{0.50} & 2.38M & 0.209 & 1.778 & \underline{0.023} \\ 
\midrule
\textbf{{\comicneue\textbf{MicroCharNet}}} & \bluecheck & \textbf{0.83} & \textbf{0.52} & \textbf{0.85} & \textbf{0.51} & \textbf{0.08M} & \textbf{0.096} & \textbf{1.023} & \textbf{0.021} \\ 
\bottomrule
\end{tabular}%
\end{table*}

At the end of the backbone, a Coordinate Attention (\textcolor{CoordAtt}{\textbf{CoordAtt}}) module is integrated to enhance spatial feature representation while maintaining low computational overhead. Unlike conventional attention mechanisms that compress the entire spatial domain into a single global descriptor, CoordAtt decomposes the contextual encoding into two separate spatial directions. This design enables the module not only to capture inter-channel dependencies but also to preserve positional information along each coordinate axis, which is particularly beneficial for detecting small and geometrically structured objects such as license plate characters. Given an input feature tensor $X_{attn} \in \mathbb{R}^{C_2 \times \frac{H}{8} \times \frac{W}{8}}$, which is produced by the preceding layers, \textcolor{CoordAtt}{\textbf{CoordAtt}} first performs two one-dimensional feature aggregation operations. Specifically, the feature map is pooled along the width direction to generate a height-aware representation $X_h \in \mathbb{R}^{C_2 \times \frac{H}{8} \times 1}$, and pooled along the height direction to generate a width-aware representation $X_w \in \mathbb{R}^{C_2 \times 1 \times \frac{W}{8}}$.

After permuting the width-aware branch to match the spatial layout of the height-aware branch, the two representations are concatenated along the spatial dimension and passed through a shared $1 \times 1$ convolution, followed by batch normalization and the hard-swish activation function. This transformation produces a compact intermediate representation that jointly encodes coordinate information from both spatial directions. The resulting feature is then split back into two branches corresponding to the height and width directions. Each branch is processed by an independent $1 \times 1$ convolution and normalized by a sigmoid activation to generate two attention maps $
A_h \in \mathbb{R}^{C_2 \times \frac{H}{8} \times 1}$ and 
$A_w \in \mathbb{R}^{C_2 \times 1 \times \frac{W}{8}}$.

Finally, these two attention maps are applied to the input feature tensor through element-wise multiplication:

\begin{equation}
Y_{attn} = X_{attn} \odot A_h \odot A_w,
\end{equation}
where $\odot$ denotes broadcasted element-wise multiplication.

This design allows the network to emphasize informative regions while preserving positional sensitivity along both spatial axes. For license plate character detection, \textcolor{CoordAtt}{\textbf{CoordAtt}} is particularly suitable because characters are typically small and densely distributed. Qualitative visualization in Figure \ref{fig:gradcam} shows that incorporating \textcolor{CoordAtt}{\textbf{CoordAtt}} enables the network to focus more effectively on character regions than the counterpart without \textcolor{CoordAtt}{\textbf{CoordAtt}}, thereby improving the preservation of fine-grained spatial structures that are crucial for both localization and classification.

\subsection{Neck}

Based on the deep features extracted by the backbone, the neck is designed to be lightweight and perform multi-level feature fusion. Specifically, the feature map after Coordinate Attention is upsampled and concatenated with a shallower intermediate feature map. This fusion mechanism allows the model to combine high-level contextual information from deeper layers with fine-grained spatial details from higher-resolution features, thereby improving the detection of small characters.

\begin{figure}[!b]
    \centering
    \includegraphics[width=\linewidth]{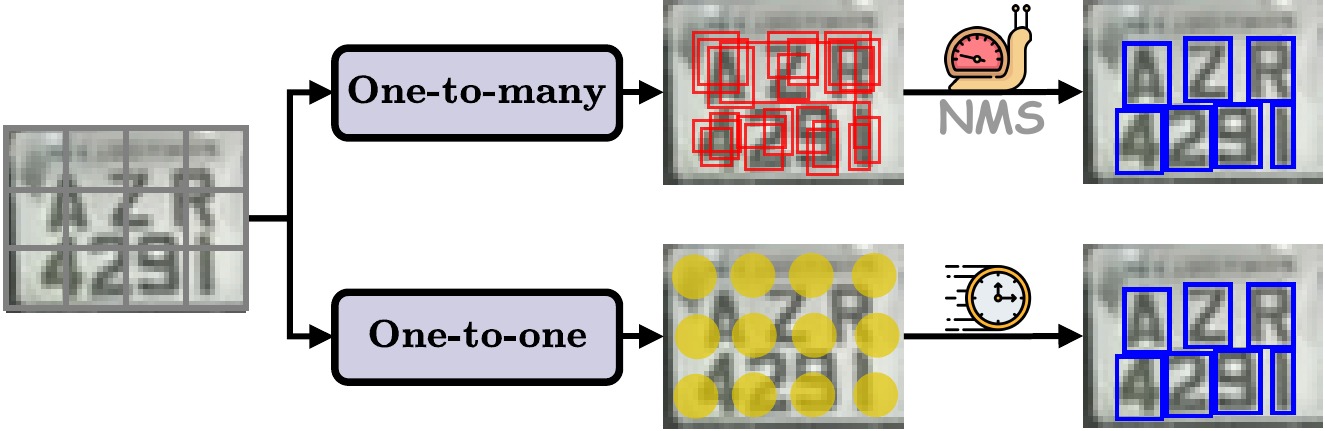}
    \caption{Comparison between conventional NMS-based inference and the proposed NMS-free inference. The one-to-many branch produces dense and redundant predictions that require {\comicneue \textcolor{nms}{\textbf{NMS}}} for duplicate removal, whereas the one-to-one branch directly generates non-redundant detections, enabling simpler and more efficient post-processing.}
    \label{fig:head}
\end{figure}

After fusion, a \textcolor{c3k2Color}{\textbf{C3k2}} block is applied to refine and reorganize the feature representation. Essentially, \textcolor{c3k2Color}{\textbf{C3k2}} is a lightweight variant derived from the \textcolor{c2fColor}{\textbf{C2f}} structure, designed to improve feature aggregation efficiency without significantly increasing computational cost. By combining short information paths with effective local transformations, \textcolor{c3k2Color}{\textbf{C3k2}} enhances the representation of fine-grained geometric details, including edges, curves, and character local structures. This is particularly beneficial for license plate character detection, where objects are small, densely arranged, and differ only in subtle geometric variations. Therefore, the proposed neck not only performs feature fusion but also plays a critical role in producing more discriminative representations for the detection head.

\subsection{Detection head}

The detection head takes the fused feature map from the neck and predicts both character locations and class labels. Instead of using a complex multi-level structure, we adopt a single-level detection head, which is suitable for license plate character detection because character scales are relatively consistent after plate cropping or normalization. This design simplifies prediction and reduces computational cost. The proposed head follows an anchor-free paradigm, where each spatial location corresponds to a candidate prediction, eliminating the need for predefined anchor boxes. In addition, the end-to-end prediction mechanism reduces intermediate post-processing steps, as illustrated in Figure~\ref{fig:head}. To further improve efficiency, we set $\texttt{reg-max}=1$, enabling direct bounding-box regression instead of distribution-based regression.

Overall, the single-level, anchor-free, and direct-regression design allows the detection head to remain compact while providing sufficient capability for license plate character detection.

\section{Results}

\subsection{Dataset and Experimental Settings}

This study focuses on license plate character detection, evaluated on the UFPR-ALPR dataset \cite{UFPR-ALPR}. The dataset contains 4,500 fully annotated images, comprising over 30,000 license plate characters collected from 150 vehicles under real-world conditions. The data are split into 1,800 training, 900 validation, and 1,800 test images. The license plates exhibit variations in color and size, and the character set includes 36 classes corresponding to alphanumeric symbols.

All experiments are conducted using the Ultralytics framework\footnote{\url{https://github.com/ultralytics/ultralytics}} with default hyperparameters unless otherwise specified. The input image resolution is set to 128$\times$128, and the model is trained for 100 epochs with a batch size of 16 using the MuSGD optimizer \cite{yolo26,muon}. All experiments are performed on a system equipped with an NVIDIA RTX 4090 GPU and an Intel(R) Core(TM) i9-10900X CPU.

\subsection{Quantitative Comparison}

To evaluate the effectiveness of {\comicneue\textbf{MicroCharNet}}, we compare it with several YOLO variants \cite{yolov8,yolov9,yolov10,yolov11,yolov12,yolo26} and the CTM model~\cite{ctm}. For a fair lightweight comparison, we use the smallest official configuration available for each YOLO variant. All models are trained within the Ultralytics framework using identical experimental settings, including the input size of $128 \times 128$, the MuSGD optimizer, and the same training hyperparameters. Inference and post-processing latency are measured on a CPU to better reflect real-world deployment on resource-constrained devices.

\begin{figure}[!t]
  \centering
  \subfloat[Successful detection examples under various conditions \label{fig:success}]{
    \includegraphics[width=0.95\columnwidth]{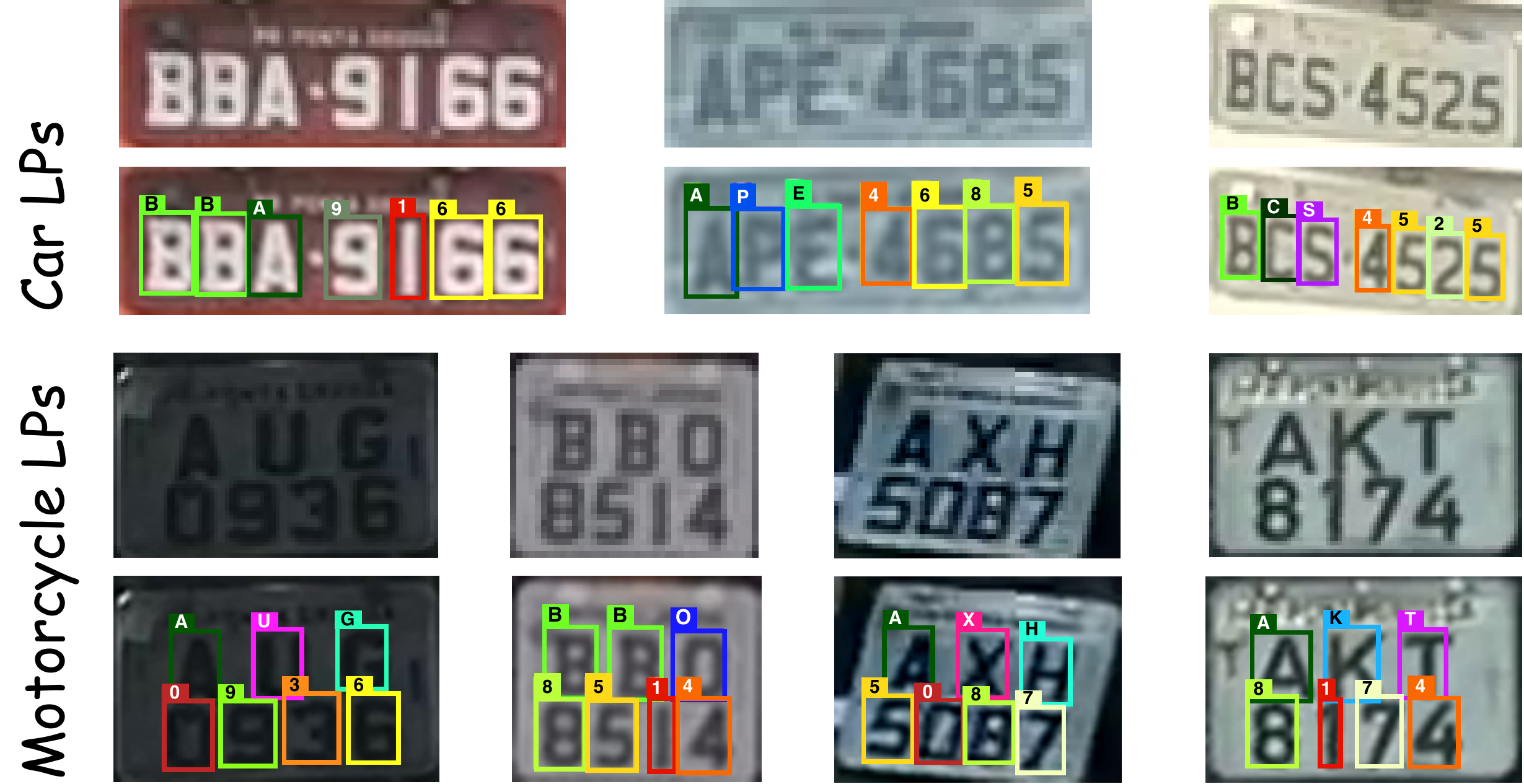}
  }\\
  
  \subfloat[Failure cases where the model struggles under challenging conditions \label{fig:limitation}]{
    \includegraphics[width=0.95\columnwidth]{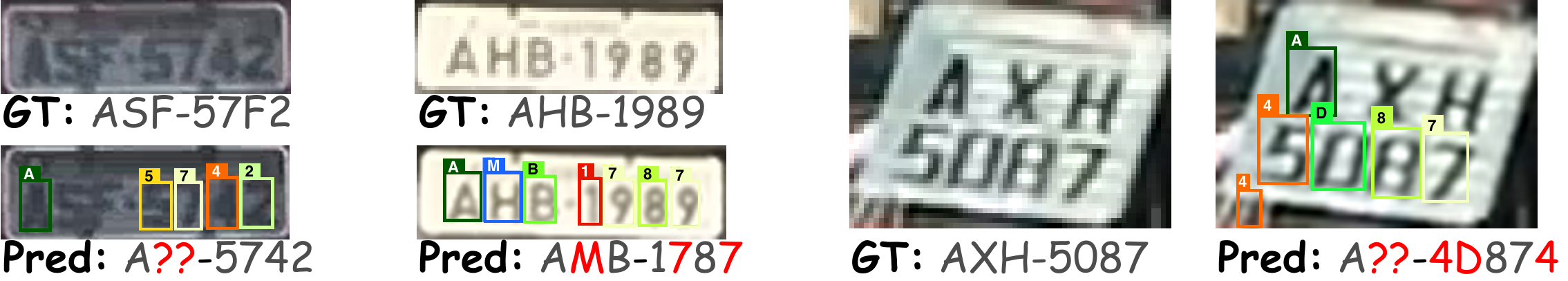}
  }
  
  \caption{Qualitative results on the UFPR-ALPR test set.}
  \label{fig:visualize}
\end{figure}

As shown in Table~\ref{tab:results}, {\comicneue\textbf{MicroCharNet}} achieves a favorable balance between detection accuracy and computational efficiency. On the validation set, the proposed model obtains the highest mAP@50 of 0.83 and reaches 0.52 mAP@50-95, which is comparable to the best-performing baseline. On the test set, {\comicneue\textbf{MicroCharNet}} achieves the highest mAP@50 of 0.85, while its mAP@50-95 of 0.51 is on par with the strongest YOLO baselines. These results indicate that the proposed lightweight architecture maintains competitive detection performance, particularly under the mAP@50 metric, while remaining robust under the stricter mAP@50-95 evaluation. In terms of efficiency, {\comicneue\textbf{MicroCharNet}} shows a clear advantage over all compared methods. It requires only 0.08M parameters and 0.096 GFLOPs, which are substantially lower than those of the YOLO variants and CTM. Despite its much smaller model size, it achieves the fastest inference time of 1.023 ms. Moreover, its post-processing time is the lowest among all methods. Compared with other NMS-free detectors such as YOLOv10 \cite{yolov10} and YOLO26 \cite{yolo26}, {\comicneue\textbf{MicroCharNet}} further reduces post-processing latency by using a single-level detection head with fewer prediction candidates.

Overall, the results demonstrate that a task-specific and lightweight architecture can not only significantly reduce model complexity but also surpass larger general-purpose detection models in both accuracy and efficiency. This confirms the effectiveness of {\comicneue\textbf{MicroCharNet}} for real-time license plate character detection on resource-constrained devices.

\subsection{Qualitative Results}

Qualitative results on the UFPR-ALPR test set are illustrated in Figure~\ref{fig:visualize}. Specifically, Figure~\ref{fig:success} presents several successful detection examples from {\comicneue\textbf{MicroCharNet}}, demonstrating its ability to accurately detect characters on both car and motorbike license plates under various conditions, including rotated plates, color variations, and different lighting environments. These results indicate that the proposed model is robust to both geometric and photometric variations commonly encountered in real-world scenarios.

In addition to successful cases, Figure~\ref{fig:limitation} shows several challenging examples where the model fails or produces suboptimal predictions. In particular, performance degrades under adverse conditions such as strong glare, blurred or low-quality license plates, and extreme rotations. These examples suggest that although the proposed model performs well in most cases, handling severe image degradation and extreme geometric variations remains a challenge for future work.

\subsection{Ablation study}

\begin{table}[!t]
\centering
\caption{Ablation study of the proposed {\comicneue\textbf{MicroCharNet}}.}
\label{table:ab}
\begin{tabular}{lcccc}
\toprule
    & \textbf{mAP@50} & \textbf{mAP@50-95} & \textbf{\#Params} & \textbf{GFLOPs} \\ \midrule
{\comicneue\textbf{MicroCharNet}}     & 0.85            & 0.51              & 82.5K             & 0.096           \\ \midrule
\textit{w/o \textcolor{CoordAtt}{CoordAtt}}     & 0.82            & 0.51              & 80.8K             & 0.095           \\
\textit{w/o \textcolor{c3k2Color}{C3k2}}        & 0.84            & 0.53              & 129.7K            & 0.193           \\
\textit{w/o \textcolor{c2fColor}{C2f}}          & 0.84            & 0.52              & 119.2K            & 0.129           \\ 
\textit{w/o end2end}      & 0.83            & 0.51              & 135.1K            & 0.203           \\
\textit{w/o single scale} & 0.87            & 0.54              & 111.1K            & 0.118           \\ \bottomrule
\end{tabular}%
\end{table}

\subsubsection{Effects of different components}
Table \ref{table:ab} shows that each component in {\comicneue\textbf{MicroCharNet}} contributes to the final trade-off between accuracy and efficiency. Removing \textcolor{CoordAtt}{\textbf{CoordAtt}} results in the largest drop in mAP@50, underscoring the importance of spatial information modeling for small-character detection. Replacing \textcolor{c3k2Color}{\textbf{C3k2}} or \textcolor{c2fColor}{\textbf{C2f}} increases both parameters and GFLOPs without yielding clear overall gains, confirming the effectiveness of the proposed lightweight feature-extraction design. Turning off the end-to-end setting also increases complexity and reduces accuracy. Although the multi-scale variant achieves slightly better detection performance, it introduces additional parameters and computation. Therefore, the proposed single-scale design offers a more favorable balance between accuracy and efficiency.

\subsubsection{Different resolutions}Table \ref{table:size} shows the effect of input resolution on the accuracy–efficiency trade-off. Increasing the input size from 96$\times$96 to 128$\times$128 improves detection performance noticeably, while further enlarging the resolution provides only marginal gains in mAP@50-95. In contrast, GFLOPs and latency consistently increase with input size, especially on CPU. These results indicate that 128$\times$128 provides the most favorable balance between detection accuracy and computational efficiency for the proposed model.

\begin{table}[]
\centering
\caption{Effect of input resolution}
\label{table:size}
\begin{tabular}{lccccc}
\toprule
\textbf{Size} & \textbf{mAP@50} & \textbf{mAP@50-95} & \textbf{GFLOPs} & \textbf{CPU (ms)} \\ \midrule
96            & 0.83            & 0.50              & 0.054           & 0.903             \\
128           & 0.85            & 0.51              & 0.096           & 1.023             \\
160           & 0.86            & 0.53              & 0.149           & 1.665             \\
192           & 0.86            & 0.54              & 0.215           & 1.953             \\
224           & 0.83            & 0.52              & 0.293           & 2.751             \\
256           & 0.82            & 0.52              & 0.383           & 3.697             \\ \bottomrule
\end{tabular}%
\end{table}

\subsection{Deployment}

To evaluate the practical applicability of {\comicneue\textbf{MicroCharNet}} in smart-camera scenarios, we deploy the model directly on embedded computing devices. Specifically, Table~\ref{tab:emb} reports the inference latency and CPU usage on Jetson Nano and Orange Pi 5 Plus. For CPU inference, the model is executed using ONNX Runtime. For hardware acceleration, TensorRT is used on the Jetson Nano GPU, while RKNN is used on the Orange Pi 5 Plus NPU.

\begin{table}[h]
\centering
\caption{Deployment performance of {\comicneue\textbf{MicroCharNet}} on embedded devices.}
\label{tab:emb}
\begin{tabular}{llcc}
\toprule
\textbf{Device} & \textbf{Runtime} & \textbf{Latency} & \textbf{CPU Usage} \\ 
\midrule
\multirow{2}{*}{Jetson Nano} 
& ONNX CPU & 21.1 ms & 100\% \\
& TensorRT GPU & 3.0 ms & 1.4\% \\
\midrule
\multirow{2}{*}{Orange Pi 5 Plus} 
& ONNX CPU & 3.9 ms & 90.0\% \\
& RKNN NPU & 5.4 ms & 12.6\% \\
\bottomrule
\end{tabular}
\end{table}

The results show that {\comicneue\textbf{MicroCharNet}} can achieve real-time inference on both devices. On the Jetson Nano, GPU acceleration with TensorRT significantly reduces latency from 21.1 ms to 3.0 ms while lowering CPU usage from 100\% to 1.4\%. On the Orange Pi 5 Plus, ONNX-based CPU inference achieves a latency of 3.9 ms. In contrast, RKNN-based NPU inference obtains 5.4 ms with much lower CPU usage. Although the NPU latency is slightly higher than that of CPU inference, it substantially reduces the CPU workload, which is beneficial for smart-camera systems that need to run multiple tasks in parallel. These results indicate that {\comicneue\textbf{MicroCharNet}} is suitable for real-time license plate character detection on embedded platforms.

\section{Discussion and conclusion}

\subsection{Limitations}

Despite its strong performance and efficiency, {\comicneue\textbf{MicroCharNet}} has several limitations. First, the model may struggle under challenging imaging conditions, such as significant lighting changes, blurriness, low-quality inputs, and highly tilted license plates. Some representative failure cases are illustrated in Figure ~\ref{fig:limitation}. Second, the evaluation was conducted on cropped license plate images, where background noise is relatively limited. As a result, the performance of {\comicneue\textbf{MicroCharNet}} on full-scene images and in unconstrained real-world environments needs further investigation.

\subsection{Conclusion}

In this paper, we proposed {\comicneue\textbf{MicroCharNet}}, an ultra-lightweight model for license plate character detection. The proposed design enables effective feature extraction for small and densely distributed characters while maintaining a highly efficient architecture. Experimental results on the UFPR-ALPR dataset demonstrate that {\comicneue\textbf{MicroCharNet}} achieves state-of-the-art performance with substantially reduced model complexity. In particular, the proposed model attains superior detection accuracy with only 0.08M parameters and 0.096 GFLOPs, while also achieving the fastest inference speed among all compared methods. In addition, experiments conducted on multiple hardware platforms further confirm the efficiency of the proposed approach in real-world deployment scenarios.
Overall, the results indicate that a carefully designed lightweight architecture can effectively balance accuracy and efficiency, and can even outperform larger general-purpose detection models. Therefore, {\comicneue\textbf{MicroCharNet}} provides a practical and effective solution for real-time license plate character detection, especially on resource-constrained devices.

\section{ACKNOWLEDGMENT}
This research is funded by University of Information Technology-Vietnam National University of Ho Chi Minh city under grant number CS4-2026-80038.

{\small
\bibliographystyle{IEEEtran}
\bibliography{ref}
}

\end{document}